\documentclass{article}
\usepackage{ijcai20}

\usepackage{times}
\usepackage{soul}
\usepackage{url}
\usepackage[hidelinks]{hyperref}
\usepackage[utf8]{inputenc}
\usepackage[small]{caption}
\usepackage{graphicx}
\usepackage{amsmath}
\usepackage{amsthm}
\usepackage{booktabs}
\usepackage{algorithm}
\usepackage{algorithmic}
\usepackage{amssymb}
\usepackage{color,xcolor}
\usepackage{multirow}
\usepackage{array}
\title{Homogeneous and Heterogeneous Relational Graph for Visible-infrared Person Re-identification}

\author{
Yujian Feng$^{1\ast}$\and
Feng Chen$^1$\footnote{These authors contributed equally to this work and should be considered co-first authors.}\and
Jian Yu$^2$\and
Yimu Ji$^{1}$\and
Fei Wu$^{1}$\and
Shangdong Liu$^{1}$\and
Xiao-Yuan Jing $^{3}$
\affiliations
$^1$Nanjing University of Posts and Telecommunications, China\\
$^2$Nanjing University Of Aeronautics And Astronautics, China\\
$^3$Wuhan University, China\\
\emails
\textit{fengyujian\_904@163.com, chenfeng1271@gmail.com.}
}

\begin{document}

\maketitle

\begin{abstract}
Visible-infrared person re-identification (VI Re-ID) aims to match person images between the visible and infrared modalities. Existing VI Re-ID methods mainly focus on extracting homogeneous structural relationships in an image, i.e. the relations between local features, while ignoring the heterogeneous correlation of local features in different modalities.
The heterogeneous structured relationship is crucial to learn effective identity representations and perform cross-modality matching. In this paper, we model the homogenous structural relationship by a modality-specific graph within individual modality and then mine the heterogeneous structural correlation
with the modality-specific graph of visible and infrared modality.
First, the homogeneous structured graph (HOSG) mines one-vs.-rest relation between an arbitrary node (local feature) and all the rest nodes within a visible or infrared image to learn effective identity representation.
Second, to find cross-modality identity-consistent correspondence, the heterogeneous graph alignment module (HGAM) further measures the relational edge strength between local node features of two modalities with
routing search way.
Third, we propose the cross-modality cross-correlation (CMCC) loss to extract the modality invariance of feature representations of  visible and infrared graphs.
CMCC computes the mutual information between modalities and expels semantic redundancy. Extensive experiments on SYSU-MM01 and RegDB datasets demonstrate that our method outperforms state-of-the-arts with a gain of 13.73\% and 9.45\% Rank1/mAP.
The code is available at https://github.com/fegnyujian/Homogeneous-and-Heterogeneous-Relational-Graph.

\end{abstract}

\section{Introduction}

Person re-identification(Re-ID) aims at matching individual pedestrian images in a query set to ones in a gallery set captured by different cameras. Most current Re-ID methods \cite{ahmed2015improved,zheng2015scalable,cheng2016person} rely on personal appearance under well visible light conditions \textit{i.e.}, Visible-visible (VV) Re-ID \cite{Beyond_triplet,hermans2017defense,zheng2016person,li2018harmonious}, which cannot be adapted well to illumination variations in real-world scenarios (\textit{e.g.}, low lighting environments at nighttime). Therefore, Visible-infrared person re-identification (VI Re-ID) \cite{AlignGAN,DRL,Tone,BDTR,DDAG} is proposed to find the corresponding infrared (or visible) images of the person captured by other spectrum cameras.

\begin{figure}[t]
\center
\includegraphics[width=0.45\textwidth]{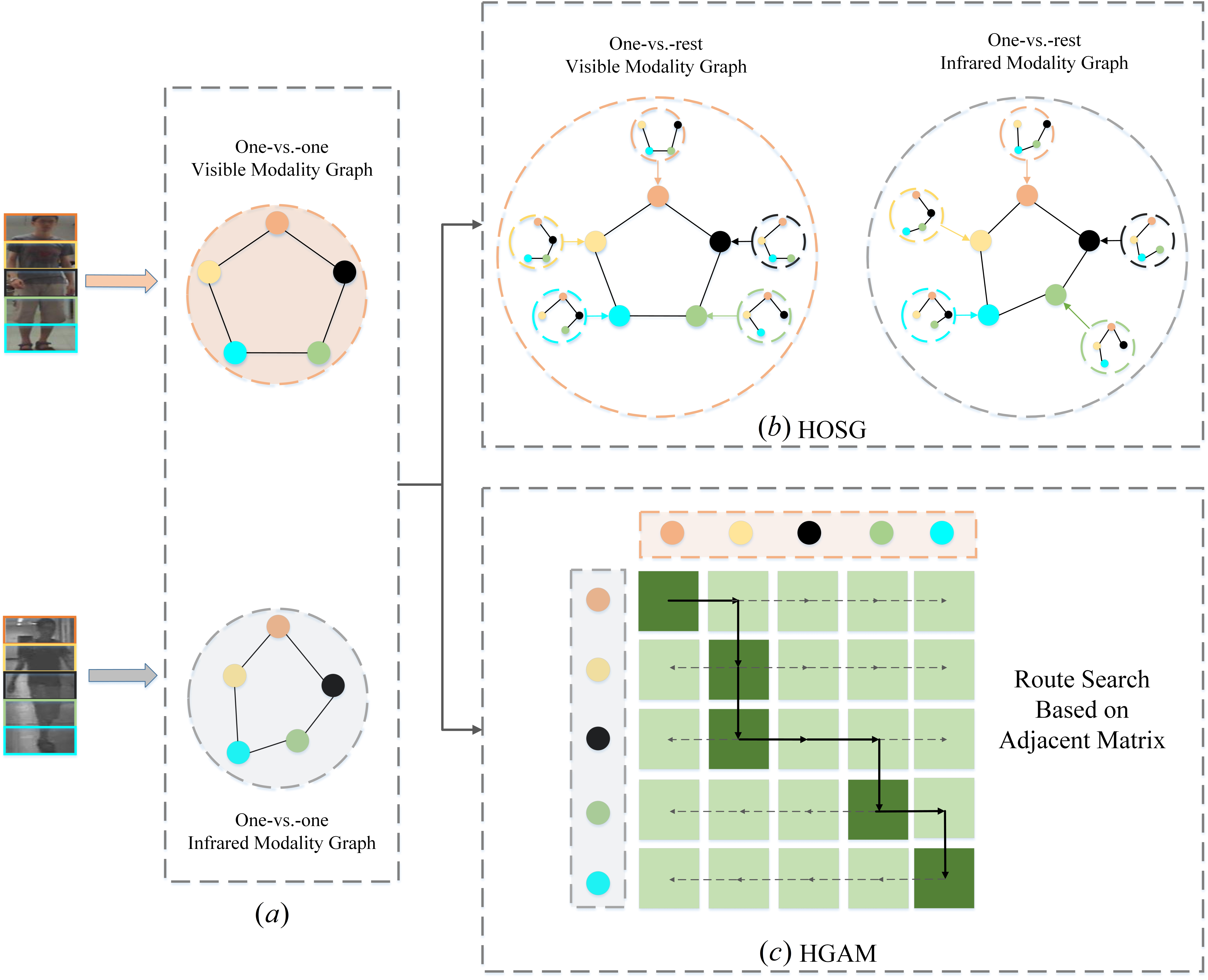}
\caption{($a$) represents the graph structural relationship between the local features of the individual modalities,
($b$) represents the visible and infrared HOSG built by local features,
and ($c$) represents HGAM constructed based on visible and infrared modality graphs.
HOSG mines one-vs.-rest  relations between nodes to explore identity representation by combining information from other local features into each local feature.
HGAM utilizes route search to align cross-modality node features by constructing a adjacent matrix.
} \label{idea}
\end{figure}


The challenges of VI Re-ID are mainly from inter-modality discrepancy resulting from intrinsically distinct imaging processes of different cameras, and intra-modality variations caused by point-of-view changes and different human postures.
Exiting methods \cite{Tone,BDTR,CGRNET,JSIA,DRL} compute the Euclidean distance relationships among images to mitigate differences between modalities. The other line works \cite{AGW,DDAG,GLMC,TMM} measure the relationship between local features from a single image to reduce intra-modality variations.
However, these two kinds of methods ignore the heterogeneous correlation between cross-modality images where each part in a visible image may correspond to a part in infrared image and vice versa.
Meanwhile, we uncover that the human parts of a person image have a strong homogeneous correlation in features for effective identity representation. The arm and head of a person in a image have a remarkably high homogeneous correlation for assembling the global representation. Therefore, it is crucial to learn the homogeneous structural relationship between intra-part features from the individual modality, and to learn the heterogeneous structural relationship of inter-part features across modalities. 




One intuitional way is to leverage the graph model which combines homogenous and heterogeneous structural relationships unitedly.
As shown in Fig. \ref{idea}, $N$ local parts (nodes) can be extracted from visible and infrared images respectively.
Therefore, the homogeneous and heterogeneous structural relationships could be reasoned in a $2N\times2N$ edge adjacent matrix. The upper-left and bottom-right sub-metrics convey the homogeneous correlation of visible and infrared modality respectively, and the rest two corners are the visible-to-infrared and infrared-to-visible correlation.
However, the computation of such a large graph is costly.
Moreover, we note that, in the homogeneous graph, node-to-node (local-to-local) relation may lack the global overview of identity representation which contributes trivially to discriminative identity representation.


In this paper, we aim to explore an effective graph model, to account for the homogeneous and heterogeneous structural relationship.
To maintain modality-specific information, we develop a homogeneous structured graph (HOSG) for an individual modality to learning identity-related representations. Different from conventional node-to-node relation which only concerns redundant local correlation, HOSG exploits one-vs.-rest relations between nodes, \textit{i.e.}, the relation between one node and all pending nodes, to reconsider the character of each node in global discriminative identity representation.
Therefore, the calculation of homogeneous edge metric and note feature updating would decrease to $2N \times 2N$.


To maintain modality-shared consistency, we propose the heterogeneous graph alignment module (HGAM) to achieve cross-modality alignment. HGAM reasons the relation between cross-modality nodes, but it is more effective.
Instead of aimlessly considering all potential cross-modality edges, we redirect it as minimum energy estimation: the relation of one node to nodes from other modality is always dominated by a specific node-to-node one.
For example, the person's head in two-modality images always has a high correlation.
Therefore, we turn to find the minimum energy of the heterogeneous edge adjacent matrix of one node as the most discriminative correlation. We search the shortest path, \textit{i.e.,} the minimum energy of sub-metric, between cross-modality nodes by adopting a route search algorithm similar to the classical Dijkstra algorithm \cite{dijkstra}. In this way, we could significantly alleviate the cross-modality discrepancy and align fine-grained local features. Note that HGAM still needs to calculate the whole metric, but the computation of node updating would decrease $N$ times. Therefore, the computation complexity of HGAM is $2 \times N \times (N+1)$.


We further enhance the global alignment of heterogeneous graphs by expelling the ambiguity of identity representation. We assume that the global graph feature of the same identity information of two modalities could reflect essential modality invariance and therefore should be identical to each other. To this end, we propose the cross-modality cross-correlation (CMCC) loss to encourage the identical representation and discourage the different representation between two graphs.

The main contributions of this paper can be summarized as follows:
\begin{itemize}
\item We design the effective HOSG and HGAM to exploit the homogeneous and heterogeneous structural relationship, which favorably facilitates the identity representation and cross-modality matching.

\item We propose the CMCC loss function to mine the invariance of global features by measure mutual information in two modality images.

\item Extensive experimental results on two standard benchmarks demonstrate that our proposed approach outperforms the state-of-the-art methods with large margins.
\end{itemize}

\section{Related Work}

\subsubsection{Visible-infrared Person Re-identification.}

VI Re-ID attempts to match visible and infrared images of a person under different cameras.
In order to alleviate the modality discrepancy, Ye \cite{Tone,BDTR} proposed global feature learning to handle the cross-modality and intra-modality variations, ensuring discriminability of feature. Li  \cite{X_modality} generated a new X-modality and exchanged information between visible modality, infrared modality, and X-modality. Lu  \cite{Cm_ssft} discussed the potential of both the modality-shared information and the modality-specific characteristics.
Wang \cite{AlignGAN}, Wang  \cite{DRL} and Dai  \cite{cmGAN} projected the features into a common space by adopting the generative adversarial training. Feng  \cite{CGRNET} presented a local modality-similarity module to reduce modality disparity while preserving identity information.
However, variations (\textit{e.g.}, human pose, perspective, background clutter, color) in the VI Re-ID task may not be taken into account.
Recently, Ye \cite{DDAG,AGW} took into account discriminative parts of interest when calculating the attentional map. Liu  \cite{TMM} partitioned a person into several horizontal local strips \cite{PCB} to extract the local part-body features. Wang \cite{JSIA} performed global set-level and fine-grained instance-level alignment. Zhao \cite{CICL} learn the color-irrelevant features to align the identity-level feature. Chen \cite{NFS} utilized neural features search to select features with relation to identity information. Nevertheless, these methods compute the similarity of local features independently result in cannot  differentiate the identities of different persons. In this paper, HOSG mines identity-related representations between individual local feature and all pending local features. HGAM aligns the local features of the two modalities on the basis of retaining the identity information of the local features.



\subsubsection{Graph Neural Network.}


\begin{figure*}[t]
\center
\includegraphics[width=0.97\textwidth]{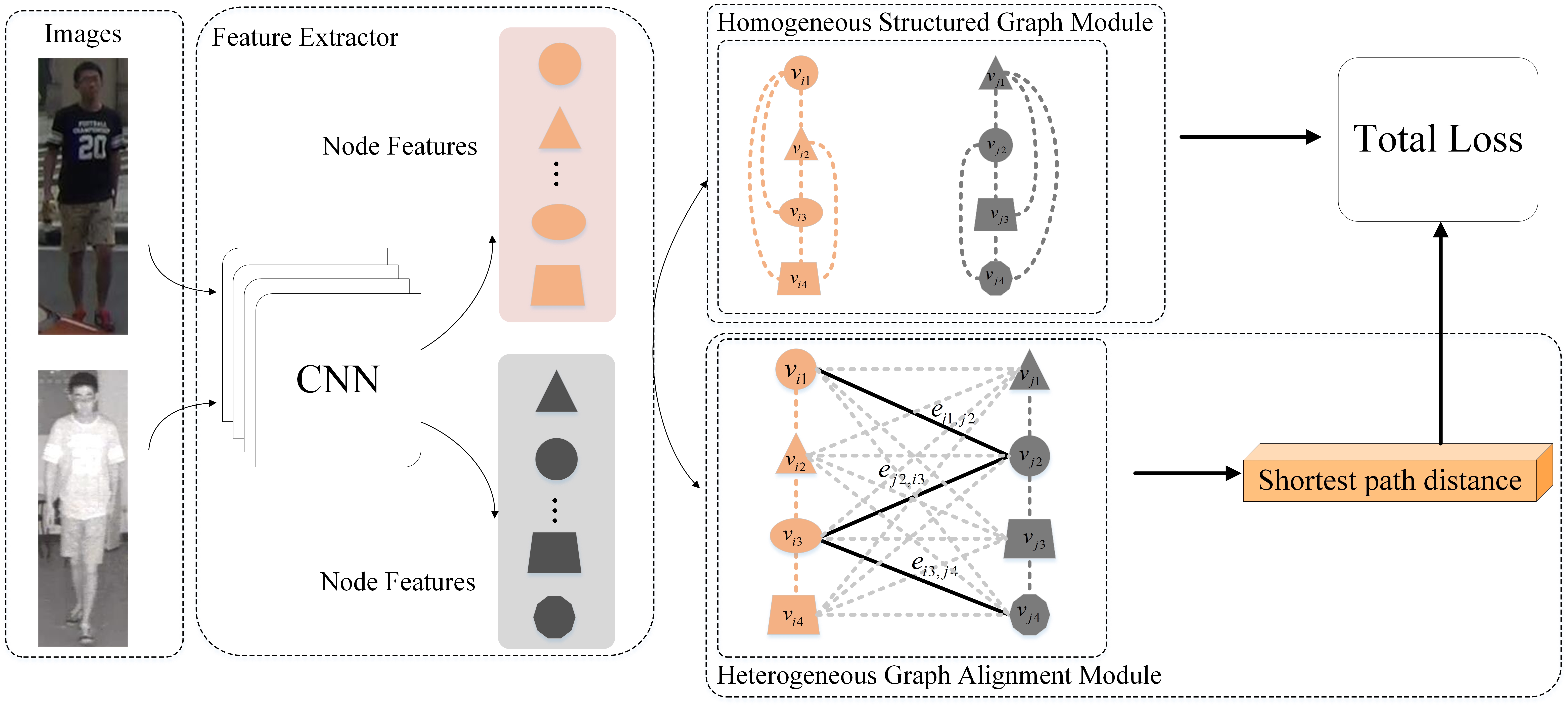}
\caption{The overall architecture or our proposed approach. We utilize a CNN backbone to extract the input visible and infrared image features. The spatial features are uniformly partitioned to obtain $H$ node features which respectively form visible and infrared homogeneous graph modules, and each node in the graph module is interconnected with other nodes.
Each homogeneous graph module constructs nodal graphs of the relationships between local features. The heterogeneous graph alignment module aligns locally similar feature nodes by searching the shortest distance among cross-modality nodes. Finally, the total loss contains $l_{id}$, $L_{tc}$ and $L_{cc}$ to guide the learning of the network.} \label{ourappraoch}
\end{figure*}

In recent years, graph convolutional network \cite{GCN_semi} and its variants \cite{GCN_heterogeneous,GCN_attention,GCN_att_net} have been successfully applied to the Re-ID task. It can generalize neural networks for structured data and transmit information between different nodes of a topology structured graph. Yan \cite{Learn_context_reid} constructed the graph model at image-level by attention module. Wu \cite{wu2020adaptive} exploited pose alignment connections and feature similarity connections to construct adaptive structure-aware neighborhood graphs. Yang \cite{spatialGCN} employed different frames of video to supplement information for each other. In the process of handing local parts relations, Zhang \cite{zhang_local_graph} aggregated the local features by attention mechanism.
However, to our knowledge, there is no method to apply graph structure to VI Re-ID tasks, and we are the first to use graph structures to construct local feature relations.



\section{Approach}

\subsection{Overview}

The overview of the our approach is illustrated in Fig. \ref{ourappraoch}. The input images including the visible and infrared images are fed into the two-stream feature extractor which is composed by ImageNet \cite{imagenet} pre-trained ResNet50 \cite{ResNet}, to extract the graph features.

The visible image graph and infrared image graph have $N_{visible}$ and $N_{ir}$ nodes respectively, and the edge between these nodes could be represented in a $N_{visible} \times N_{ir}$ matrix. As shown in Fig. \ref{idea}, the upper left and lower right sub-matrix are for the study of intra-modal autocorrelation, in order to express a better representation within the current modality. The remaining two corners are to study inter-modality correlation, which can effectively serve for cross-modality retrieval.

Specifically, we construct homogeneous graph module (HOSG) includes visible homogeneous graph and infrared homogeneous graph to look for the homogeneous relationship between the various parts of two modalities respectively.
The HOSG learns an effective feature representation of discrimination  by reasoning the relationship between individual parts of the same image.
The heterogeneous graph alignment module (HGAM) imposes the cross-modality partial relation, aiming to promote effective alignment of local part features.
Concurrently, the CMCC loss function is presented to adapt the mutual information between modalities to extract the invariance of modality feature.
These three components are integrated into a unified framework that can facilitate each other.

\subsection{Homogeneous Structured Graph Module}

The homogeneous structured module (HOSG) extracts modality-specific features containing identity information represented by visible and infrared sensor processing.
After obtaining the features from feature extractor, we conducts uniform partition on the conv-layer for learning local features \cite{PCB,relation} in terms of height dimensions. For the spatial features of each image, we divide it uniformly in the height dimension to form $N$ nodes.


For visible or infrared modality, let $G(V, E)$ denote the homogeneous structured graph of $V$ nodes where nodes ${{v}_{i}}\in V$ and edges ${{e}_{ij}}=({{v}_{i}},{{v}_{j}})\in E$.
The initially linkage weights of the edges between nodes are the L2 distance, as follows:
\begin{equation}\label{eq1}
  {{e}_{i,j}}=||f({{v}_{i}})-f({{v}_{j}})||_{2}^{2}
\end{equation}
where the $f({{v}_{i}})$ and $f({{v}_{j}})$ are represented the $i$-th and $j$-th node features, and $i,j\in [1,N]$.

The $e_{ij}$ expresses the relationship between arbitrary nodes $v_{i}$ and $v_{j}$. However, the information contained in individual node pair is not enough to investigate the global features in the modeling process. Thus, we firstly utilize one local node $v_{i}$ and all the remaining nodes $v_{r}$ for correlation determination, \textit{i.e.}, one-vs.-rest as shown in follows:
\begin{equation}\label{eq2}
  {{e}_{ir}}=(v_i,v_r)=({{v}_{i}},\sum\limits_{j\ne i}({{v}_{j}})) \approx \sum_{j\neq i} e_{ij},
\end{equation}
where $v_{r}$ is obtained by the information aggregation of the remaining nodes as Eq. \ref{eq3}.
\begin{equation}\label{eq3}
  {{f}}({{v}_{r}})=\frac{1}{N-1}\sum\limits_{i\ne j}{{{f}}({{v}_{j}})}.
\end{equation}

\begin{figure}[t]
\center
\includegraphics[width=0.4\textwidth]{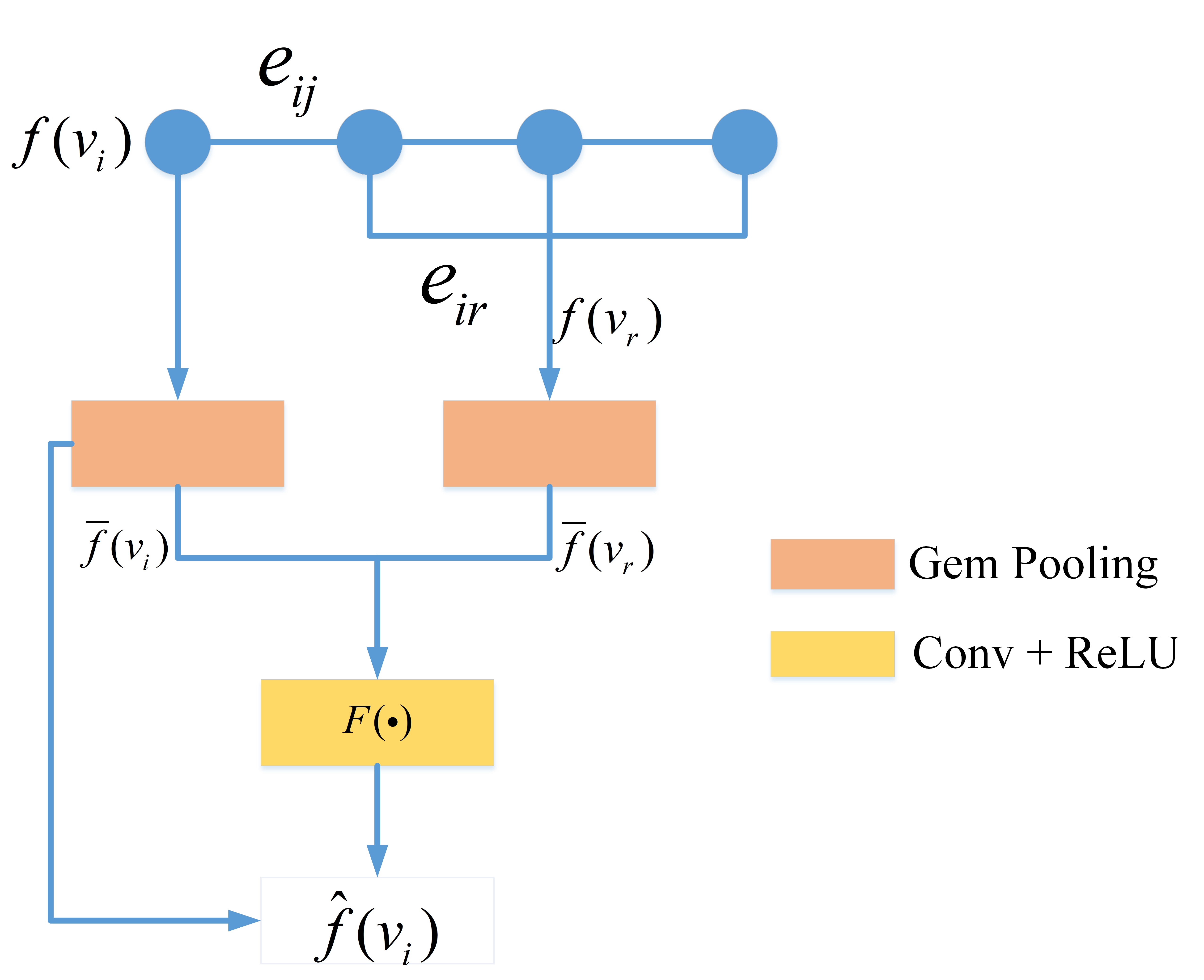}
\caption{The detailed pipeline of the HOSG. Each node is linked to each other and the weights of the edges are updated by aggregating the information of the remaining nodes.  } \label{relation}
\end{figure}

Secondly, for the updating of node feature as illustrated in Fig. \ref{relation}, we add the Gem Pooling layer separately for each $f({{v}_{i}})$ and ${{f}}({{v}_{r}})$ to obtain the intermediate features $\bar f({{v}_{i}})$ and $\bar {{f}}({{v}_{r}})$. Then, $\bar f({{v}_{i}})$ and $\bar {{f}}({{v}_{r}})$ are fused and summed up with $\bar f({{v}_{i}})$  by convolution layer to get features $\hat f({{v}_{i}})$  with information of other nodes. We utilize skip-connection form the ideology of ResNet \cite{ResNet} to transfer the relational features, as follows:
\begin{equation}\label{}
  \hat f({{v}_{i}})=\bar f({{v}_{i}})+F(\bar f({{v}_{i}}),\bar {{f}}({{v}_{r}})),
\end{equation}
where the residual $F(\cdot)$ is composed of a 1 $\times$ 1 convolution, a batch normalization and a ReLU layers sequentially. This update way takes into account the correlation with other nodes, and makes it more discriminative while preserving a compact feature representation for personal identity.


\subsection{Heterogeneous Graph Alignment Module}

To avoid the interference of noise information between two modalities images, we utilize the heterogeneous structural relationship to find the most obvious correlation among cross-modality nodes.
For example, compared with other body parts, the person head in visible and infrared images may have an intuitional higher correspondence. This setting would also remove alignment redundancy.


We construct the heterogeneous graph alignment module (HGAM) to mine the most discriminative link $e_{i,j}=(v_i,v_j)$ where $v_i\in V_{visible}$ and $v_j\in V_{ir}$, \textit{i.e.}, the minimum energy of heterogeneous adjacent matrix. Specifically, after obtaining the heterogeneous adjacent matrix by measuring the normalized Euclidean distance between node features, we aim to search the end-to-end path as shown in Fig. \ref{ourappraoch}. We regard the heterogeneous local feature alignment process as the process of node route search.
Taking arbitrary node as beginning, we search the shortest path toward heterogeneous graph. Then the arrival node would act as beginning to repeat this step until all nodes of two graphs are traversed.
This shortest path searching is similar to the Dijkstra \cite{dijkstra} algorithm and could use other flexible alternatives.

For perspicuity, we describe it in the illustration of adjacent metric, as shown in the lower left sub-metric $A_{i2v}$ of Fig. \ref{idea}. We define the local distance between the two graphs as the total distance of the shortest path from $(1, 1)$ to $(H, H)$ in the matrix $A_{i2v}$. The distance can be calculated through dynamic programming is similar to the Dijkstra \cite{dijkstra,Alignedreid}:
\begin{equation}\label{}
S_{i, j}= \begin{cases}e_{i, j} & i=1, j=1 \\ S_{i-1, j}+e_{i, j} & i \neq 1, j=1 \\ S_{i, j-1}+e_{i, j} & i=1, j \neq 1 \\ \min \left(S_{i-1, j}, S_{i, j-1}\right)+e_{i, j} & i \neq 1, j \neq 1\end{cases}
\end{equation}
where $S_{i,j}$ is the total distance of the shortest path when walking from (1, 1) to ($i$, $j$) in the adjacent metric $A_{i2v}$.
$S_{H,H}$ is the total distance of the final shortest path (\textit{i.e.}, the local distance) between the two graphs.



This shortest path searching algorithm, which calculates the correlation between local features from two images within class, can effectively perform feature alignment between different images of the same person, thus avoid the interference of noise information between two images.

\subsection{Loss Function}

We assume that the person image is a concrete representation of identity essence in real world, influenced by some actual factors, \textit{i.e.}, pose, occlusion, sensor \textit{etc}. Therefore, the ReID task could be reconsidered as recovering identity essence from cross-modality images for following retrieve. In this case, we believe the global graph feature could be deemed as such a identity representation and then expect heterogeneous global graph features to be identical. Therefore, we propose the CMCC loss function to exploit the essential relationship  between modalities.

We utilize a cross-correlation matrix $X$, which is obtained from the inherent attribute mutual information of global features. It is computed as follow:
\begin{equation}\label{eq6}
  {{X}_{ii}}=f_{visible}^{T}({{x}_{i}}){{f}_{infrared}}({{x}_{i}}),
\end{equation}
where $i$ is the $i$-th identity information, and $T$ represents the transpose of the matrix. ${{f}_{visible}}(\cdot )$ and ${{f}_{infrared}}(\cdot )$ represent the global features of HOSG of two modalities respectively.
${{X}_{ii}}$ is cross-correlation matrix between modalities, which is the consistency of the representation of the global features of the graph in indentity retrieval.

The CMCC loss $L_{cc}$ optimizes the inter-modality relationship, as follows:
\begin{equation}\label{Lcc}
  {{L}_{cc}}=\sum\limits_{i}^{M}{(1-{{X}_{ii}})^{2}}+\alpha \sum\limits_{i}^{M}{\sum\limits_{j,j\ne i}^{M}{{{X}_{ij}}^{2}}},
\end{equation}
where $M$ is the number of identity information, $\alpha$ is a positive co-efficient factor.
By minimizing $L_{cc}$, it makes ${{X}_{ii}}$ close to 1 and ${{X}_{ij}}$ close to 0.
${{X}_{ii}}\to 1$ enhances the correlation between features of two modalities within the same class, which improves the representativeness of the features.
${{X}_{ij}}\to 0$ means reducing the correlation between the $i$-th and $j$-th class, thus decreases the redundant information in graph representation.


During the training stage, we utilize three types of loss functions to constrain HOSG and HGAM, including the identity loss $L_{id}$, triplet-center loss $L_{tc}$ \cite{TMM} and cross correlation loss $L_{cc}$. The identity loss $L_{id}$ encourages an identity-invariant feature representation, and the triplet-center loss $L_{tc}$ optimizes the triplet-wise relationships among different images between modalities. Thus the final total loss can be represented as:
\begin{equation}\label{total_loss}
  {{L}_{total}}={{L}_{id}}+\beta L_{tc}^{{}}+ \gamma {{L}_{cc}}
\end{equation}
where $\beta$ and $\gamma$ are the balance coefficients and $L_{id}$ uses the cross-entropy loss.

\section{Experimental Results}
\subsection{Dataset and Evaluation Protocol}

\subsubsection{Dataset}
We evaluate our approach on two benchmarks including SYSU-MM01 \cite{SYSU} and RegDB \cite{RegDB}.

(1) SYSU-MM01 is a large-scale dataset that contains images captured by four visible and two near infrared cameras, including indoor and outdoor environments. It consists of 30,071 visible images and 15,792 NIR images of 491 identities.
The dataset was split into a training set and a testing set, in which 22,258 visible images and 11,909 infrared images of 395 identities are used for training set. The 3,803 infrared images of 96 identities and 301 randomly sampled visible images from 96 identities are used for query set and gallery set.

(2) RegDB dataset is constructed by a couple of cameras (one visible and one infrared). It contains 8,240 images of 412 identities. Each identity has 10 images from the visible camera and 10 images from the infrared one. The dataset is randomly equally split into a training dataset and testing dataset.
Following \cite{Tone,BDTR}, we evaluate both visible-to-infrared and infrared-to-visible modes, by alternatively using all visible/infrared images as the gallery set.

\subsubsection{Evaluation Protocols}
All experiments follow standard evaluation protocols including cumulative matching characteristic (CMC) and mean average precision (mAP).
Following \cite{SYSU,Tone}, we utilize both indoor-search and all-search mode evaluation protocol, which is employed by all comparison methods \cite{AGW,DDAG,Cm_ssft,CICL}. The gallery set in the all-search mode contains images from all visible cameras, while the gallery set in the indoor-search mode only contains images from two indoor visible cameras.
The results on RegDB are based on the average of 10 random splits of the training and testing sets.

\subsubsection{Implementation details}
The proposed approach is implemented with PyTorch and trained on a NVIDIA 3090 GPU.
Following the existing VI Re-ID methods \cite{BDTR,AGW,DDAG,NFS,CICL}, the ResNet50 \cite{ResNet} model pre-trained on ImageNet \cite{imagenet} is adopted as our backbone network.
In the training phase, the visible and infrared images are resized to 288 $\times$ 144.
For each mini-batch, a total of 4 identities are randomly chosen, each of which is randomly sampled with 8 visible images and 8 infrared images.
We adopt the stochastic gradient descent (SGD) to optimize, and then train the model with a total of 60 epochs. The learning rate is initialized to 0.01 and decays 10 times each ten epoch.
The $N$ in Eq. \ref{eq3} is set to 6 for two modalities, and the hype-parameter $\beta$ in Eq. \ref{total_loss} is set to 2  \cite{TMM} for both SYSU-MM01 and RegDB datasets.

\subsection{Comparison with State-of-the-art Methods}

\begin{table*}[t]\caption{Comparison with state-of-the-arts on single-shot SYSU-MM01 where ‘R*’ denotes Rank*. The mAP denotes mean average precision score (\%).}
\setlength{\belowcaptionskip}{-1cm}
\centering
\setlength{\tabcolsep}{2.5mm}{
\begin{tabular}{c|c|c|c|c|c|c|c|c|c}
\hline
 \multirow{2}{*}{Method} & \multirow{2}{*}{Source} & \multicolumn{4}{c|}{All-Search} & \multicolumn{4}{c}{Indoor-Search} \\ \cline{3-10}
 &                           & R1     & R10    & R20    & mAP   & R1     & R10     & R20    & mAP   \\ \hline
TONE       & AAAI-2018       & 12.52  & 50.72   & 68.60   & 14.42  & -      & -       & -       & -     \\ \hline
BDTR       & IJCAI-2018      & 17.01  & 55.43   & 71.96    & 19.66  & -      & -       & -       & -     \\ \hline
D$^2$RL   & CVPR-2019        & 28.90  & 70.60   & 82.40   & 29.20  & -      & -       & -       & -     \\ \hline
JSIA       & AAAI-2020       & 38.10  & 80.70   & 89.90   & 36.90  & 43.80   & 86.20    & 94.20    & 52.90  \\ \hline
AlignGAN   & ICCV-2019       & 42.40  & 85.00  & 93.70   & 40.70  & 45.90   & 87.60    & 94.40    & 54.30  \\ \hline
CMSP        & IJCV-2020      & 43.56  & 86.25  & -      & 44.98 & 48.62  &89.50    & -       & 57.50  \\ \hline
AGW         & TPAMI-2021     & 47.50  & -      & -      & 47.65 & 54.17  &-       & -       & 62.97  \\ \hline
Xmodel      & AAAI-2020      & 49.92  & 89.79  & 95.96  & 50.73 & -      &-       & -       & -     \\ \hline
DDAG       & ECCV-2020       & 54.75  & 90.39  & 95.81  & 53.02 & 61.02  &94.06   & 98.41   & 67.98     \\ \hline
NFS         & CVPR-2021      & 56.91  & 91.34  & 96.52  & 55.45 & 62.79  &96.53   & 99.07   & 69.79     \\ \hline
CICL        & AAAI-2021      & 57.20  & 94.30  & \textbf{98.40}   & 59.30  & 66.60   & \textbf{98.80}    & 99.70    & 74.70     \\ \hline
cm-SSFT     & CVPR-2020      & 61.60  & 89.20  &93.90   & 63.20   & 70.50  & 94.90    &97.70     & 72.60        \\ \hline
HCT         & TMM-2020       & 61.68  & 93.10  & 97.17  & 57.51 & 63.41  &91.69   & 95.28   & 68.17     \\ \hline
GLMC     & TNNLS-2021        & 64.37  & 93.90  & 97.53  & 63.43 & 67.35  &98.10   & \textbf{99.77}   & 74.02     \\ \hline
Ours       &                & \textbf{78.10} & \textbf{94.50}  & 97.27  &\textbf{72.88} & \textbf{81.70}  & 95.02   & 98.37   & \textbf{81.44} \\ \hline
\end{tabular}}\label{table sysu}
\end{table*}

We compare the proposed with state-of-the-art VI Re-ID methods,
including \textbf{metric learning based methods} (TONE \cite{Tone}, BDTR \cite{BDTR});
\textbf{GAN-based methods} (Xmodal\cite{X_modality}, D$^2$RL \cite{DRL}, JSIA\cite{JSIA}, AlignGAN \cite{AlignGAN});
\textbf{local features learning based methods} (AGW \cite{AGW}, DDAG \cite{DDAG}, HCT \cite{TMM}, GLMC \cite{GLMC});
\textbf{shared and specific feature learning based methods} (cm-SSFT \cite{Cm_ssft}, CMSP \cite{CMSP}, NFS \cite{NFS}, CICL \cite{CICL}).

\subsubsection{Comparisons on SYSU-MM01.}

\begin{figure}[h]
\centering{
\includegraphics[width=0.45\textwidth]{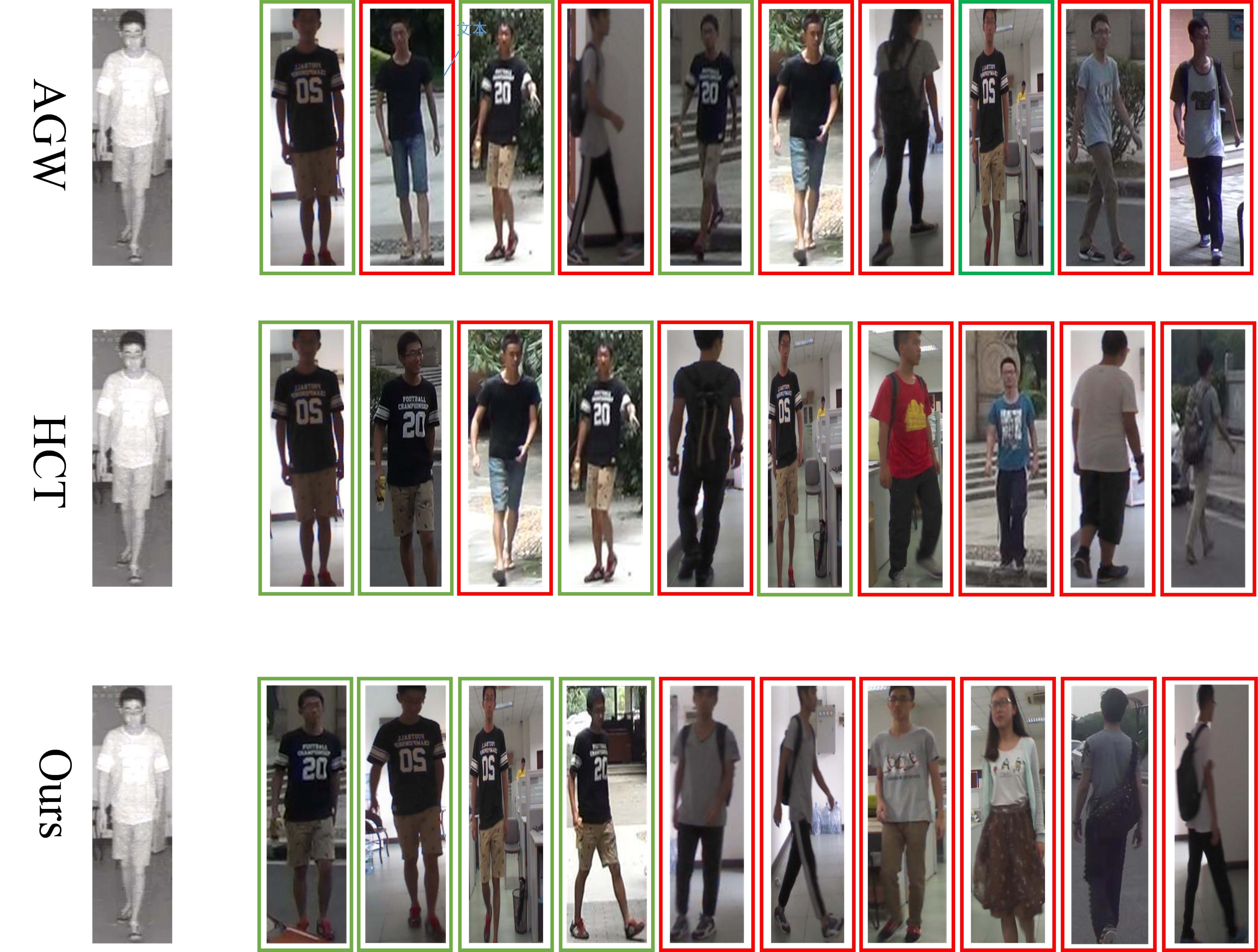}
}
\caption{Rank-10 results of AGW and HCT methods, and our approach on SYSU-MM01 dataset in all-search mode. The green and red rectangle represent true match and false match, respectively. According to the description of the SYSU-MM01 dataset, each infrared query can have four true matches at most.} \label{test_search}
\end{figure}

The comparison results on SYSU-MM01 dataset are shown in Table \ref{table sysu}.
Our approach is largely superior to the existing methods \cite{GLMC,Cm_ssft}
In the most challenging all-search mode, our approach achieves 78.10\% Rank-1 and 72.88\% mAP, which significantly outperforms the GLMC \cite{GLMC} by 13.73\% in Rank-1 accuracy and 9.45\% in mAP.
In indoor-search mode, our method shows a significant improvement over the cm-SSFT \cite{Cm_ssft}, with an 11.20\% increase in Rank-1 accuracy and an 8.84\% increase in mAP.
In order to visualize the query results, the randomly selected several queries is presented in Fig. \ref{test_search} on SYSU-MM01 dataset under in all-search mode. The result of matching demonstrates the effectiveness of our approach.

\subsubsection{Comparisons on RegDB}
The results of the RegDB dataset are listed in Table \ref{table regdb}.
Compared to GLMC \cite{GLMC}, we achieve Rank-1 accuracy of 94.92\% and mAP of 94.58\%, significantly improving the Rank-1 by 3.08\% and mAP by 13.16\% under the visible-infrared mode. Our approach far outperforms to existing methods under the infrared-visible mode, achieving Rank-1 accuracy of 93.35\% and mAP of 93.98\%, with a 2.23\% improvement in Rank-1 and a 12.92\% improvement in mAP.

\begin{table}[h]\caption{Comparison with the state-of-the-arts on the RegDB dataset. V - I means visible-search-infrared mode, while I - V means infrared-search-visible mode.}
\setlength{\belowcaptionskip}{0.3cm}
\centering
\setlength{\tabcolsep}{1.0mm}{
\begin{tabular}{c|c|c|c|c|c}
\hline
\multirow{2}{*}{Method} & \multirow{2}{*}{Source} & \multicolumn{2}{c|}{V - I} & \multicolumn{2}{c}{I - V} \\ \cline{3-6}
              &        & R1      & mAP     & R1        & mAP   \\ \hline
TONE     & AAAI-2018  & 16.87  & 14.92   & 13.86    & 16.98 \\ \hline
BDTR     & IJCAI-2018 & 33.47  & 31.83   & 32.72    & 31.10 \\ \hline
D$^2$RL  & CVPR-2019  & 43.40   & 44.10    & -      & - \\ \hline
JSIA     & AAAI-2020  & 48.50   & 49.30    & 48.10   & 48.90  \\ \hline

AlignGAN & ICCV-2019  & 57.90    & 53.60    & 56.30  & 53.40  \\ \hline
Xmodel   & AAAI-2020  & 62.21  & 60.18   & -        & -     \\ \hline
CMSP     & IJCV-2020  & 65.07  & 64.50     & -        & -     \\ \hline
DDAG     & ECCV-2020  & 69.34  & 63.46    & 68.06    & 61.80       \\ \hline
AGW      & TPAMI-2021 & 70.05   & 66.37    & -       & -  \\ \hline
cm-SSFT  & CVPR-2020  & 72.30    & 72.90   & 71.00   & 71.70  \\ \hline
CICL     & AAAI-2021  & 78.80   & 69.40    & 77.90     & 69.40  \\ \hline
NFS      & CVPR-2021  & 80.54  & 72.10    & 77.95     & 69.79  \\ \hline
HCT      & TMM-2020   & 91.05  & 83.28    & 89.30     & 81.46  \\ \hline
GLMC     & TNNLS-2021 & 91.84  & 81.42    & 91.12     & 81.06  \\ \hline
Ours      &            & \textbf{94.92}  & \textbf{94.58}    &\textbf{93.35}        &  \textbf{93.98}  \\ \hline
\end{tabular}}\label{table regdb}
\end{table}

\subsection{Ablation Study}

\begin{table}[h]\caption{Evaluation of each component on SYSU-MM01 dataset.}
\centering
\setlength{\tabcolsep}{3mm}
\begin{tabular}{c|c|c|c|c}
\hline
 \multirow{2}{*}{Setting}& \multicolumn{2}{c|}{All-Search} & \multicolumn{2}{c}{Indoor-Search} \\ \cline{2-5}
                     & R1     & mAP     & R1        & mAP   \\ \hline

\textit{B}           & 61.68    & 57.51 & 63.41    & 68.17 \\\hline
\textit{B} + HOSG    & 72.00   & 67.44  &72.37     &75.14  \\\hline
\textit{B} + HGAM    & 72.92    & 69.67 & 77.35    &78.06 \\\hline
\textit{B} + CMCC    & 71.52   & 68.52 &75.41     &77.82 \\\hline
\end{tabular}\label{table component}
\end{table}

To further demonstrate the effectiveness of the HOSG, HGAM and CMCC, the proposed approach is evaluated under three different settings for SYSU-MM01 dataset, \textit{i.e.} with or without HOSG, HGAM and CMCC.
The \textit{B} denotes the baseline of ResNet-50 as the backbone network with learning objective $L_{hc}$ \cite{TMM}.
To illustrate the contribution of each module or objective function, we add them into the model one by one.

\subsubsection{The Effectiveness of HOSG,  HGAM and CMCC}

As shown in Table \ref{table component},
we utilize HOSG module to improve the Rank-1 accuracy by 10.32\% and 8.96\% in both all-search and indoor-search modes compared with \textit{B}, respectively.
It can be observed that using the HGAM module clearly brings 11.24\% Rank-1 and 3.3\% mAP increases in all-search mode liken to \textit{B}. It significantly improves 13.94\% Rank-1 and 9.89\% mAP in indoor-search mode.
Moreover, the effect of using CMCC loss improves 9.84\% Rank-1 and 2.51\% mAP in the all-search mode, and improves 12.00\% Rank-1 and 9.65\% mAP in the indoor-search mode.
The above results for the SYSU-MM01 dataset demonstrate the outstanding performance of our approach.
This is mainly due to HOSG mining the correlation between local features of individual modality and HGAM effectively aligning the local features between modalities. Meanwhile, the utilization of CMCC makes the global features between modalities more remarkable and avoids the interference of noise information.

\subsubsection{The Influence of the Hype-parameter $\alpha$ and $\gamma$}

\begin{figure*}[htp]
\centering{
\includegraphics[width=1\textwidth]{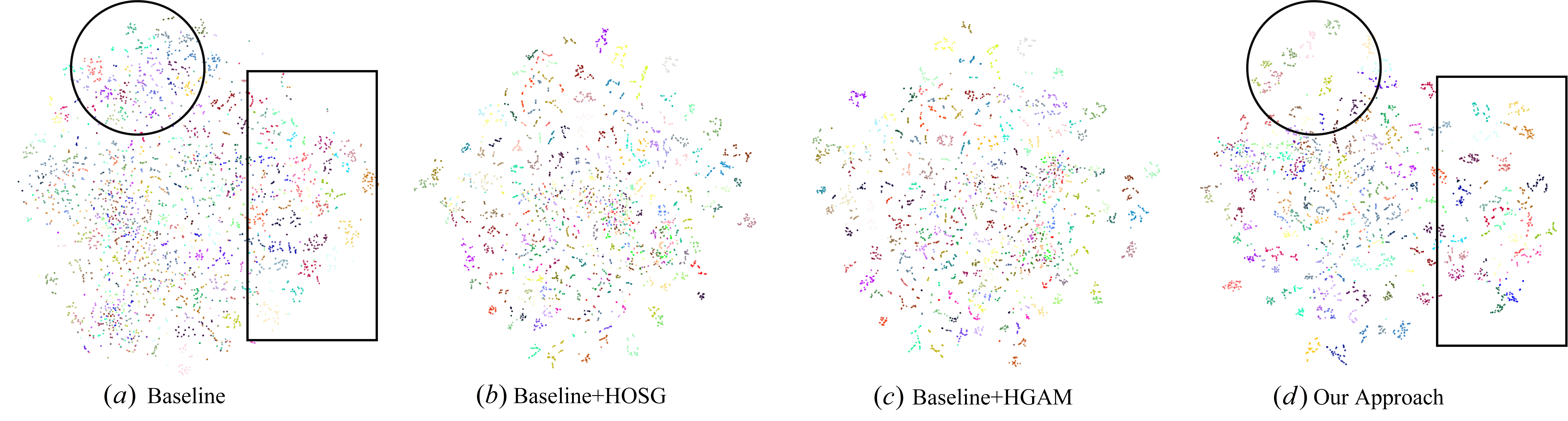}
}
\caption{ T-SNE visualization of the learned feature distributions of our approach and baseline methods on SYSU-MM01 dataset under in all-search mode.} \label{visiual}
\end{figure*}

\begin{figure}[htp]
\centering{
\includegraphics[width=0.48\textwidth]{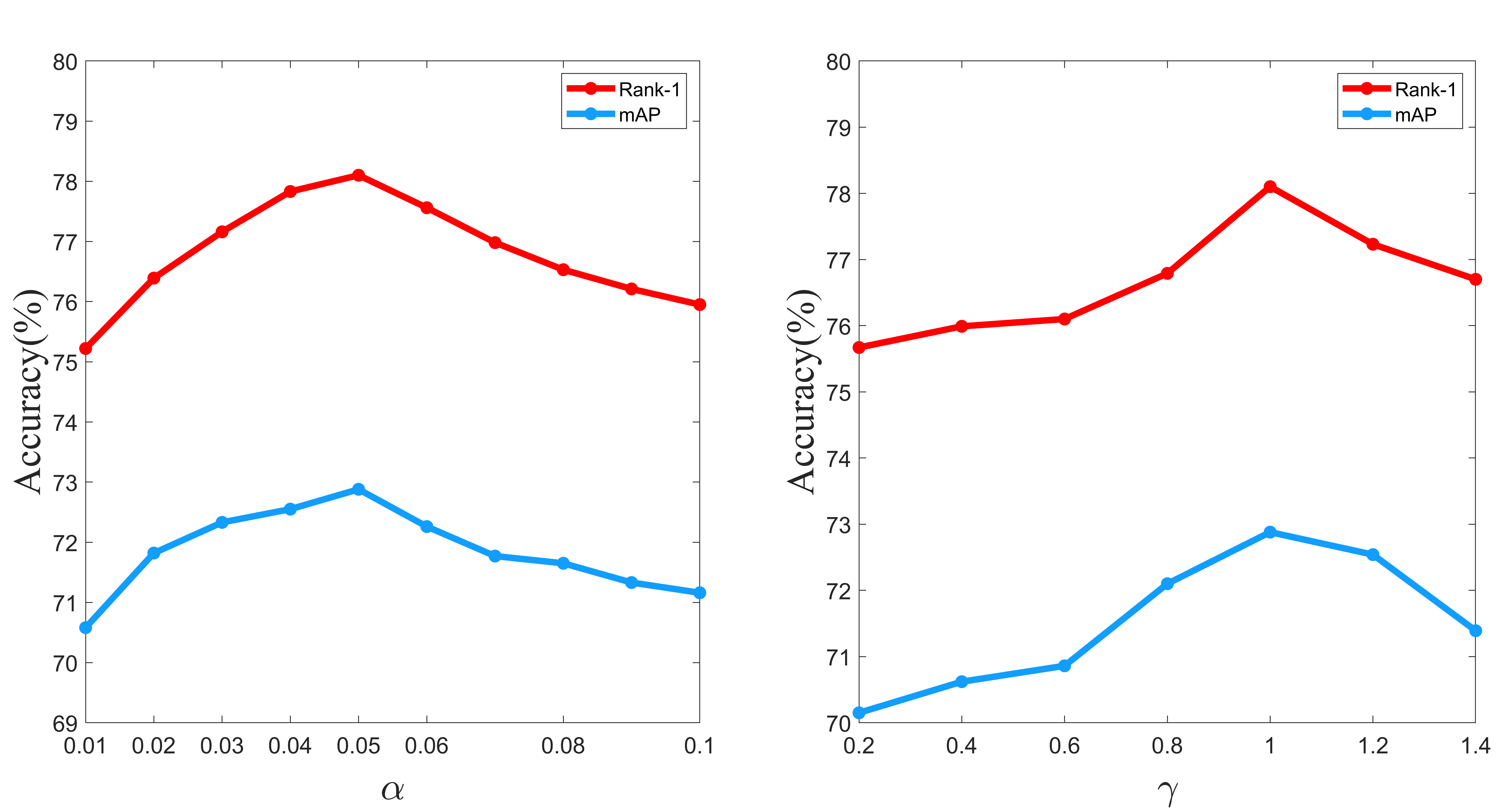}
}
\caption{The effect of Rank-1 and mAP when $\alpha$ and $\gamma$ take different values under all-search mode on the SYSU-MM01 dataset.} \label{lambda}
\end{figure}

By adjusting different balance paraments, the proposed approach is evaluated with the different value of $\alpha$ and $\gamma$ on the SYSU-MM01 dataset under the all-search mode.
As shown in Fig. \ref{lambda} (left), we set the $\alpha$ from 0.01 to 0.1 to perform the verification of the effect of our approach. The best results are obtained when $\alpha$ takes the value of 0.05.
Moreover, as shown in Fig. \ref{lambda} (right), the trade-off coefficient $\gamma$ is increased from 0.2 to 1.4 with constant improvements of the results, and the best Rnak-1 accuracy is obtained when $\gamma$ = 1.0.

\subsection{Visualization of Learned Features}

In order to further verify the effectiveness of our approach, we utilize t-SNE \cite{tsne} to visualize the features in 2D plane on SYSU-MM01 dataset. As shown in Fig. \ref{visiual}, (a - d) represent the visualization of the baseline, baseline + HOSG, baseline + HGAM, and our approach.
The different colors indicate different categories, the circle and rectangle represent the distribution change of the feature.

We observed three phenomenons. (1) The intra-class distribution in the circle part of Fig. \ref{visiual} (a) is scattered, and the features of the same identity are not well clustered together, while the features between identities are not separated. (2) Compared to Fig. \ref{visiual} (a), HOSG and HGAM modules are able to aggregate the features of the same identity information well. (3) Our approach can effectively aggregate features of the same identity information together, and makes the features of different identities far away from each other, thus improves the feature discrimination.

\section{Conclusion}

In this paper, we have proposed HOSG and HGAM to model homogeneous and heterogeneous structural relationships in a unified framework for VI Re-ID by constructing the visible and infrared graph structures. Specially, HOSG exploits one-vs.-rest relations between nodes to reconsider the feature representation of each node in the global discriminative identity representation. HGAM aligns the local features of the two modalities by adopting route search algorithm. Moreover, CMCC measures mutual information in two modality images enhance the invariance of global features and thus reduce the redundant information of features. Extensive experiments validate the effectiveness of our proposed approach.


\bibliographystyle{unsrt}
\bibliography{ijcai20}

\end{document}